# Analysis of Feature Detector and Descriptor Combinations with a Localization Experiment for Various Performance Metrics

**Ertugrul BAYRAKTAR*, Pınar BOYRAZ**

*Graduate School of Science Engineering and Technology Mechatronics Engineering Department, Istanbul Technical University,* bayraktare@itu.edu.tr

*Mechanical Engineering Department, Istanbul Technical University, Inonu Cd. No: 65, 34437 Beyoglu Istanbul, Turkey, pboyraz@itu.edu.tr*

## Abstract

The purpose of this study is to give a detailed performance comparison about the feature detector and descriptor methods, particularly when their various combinations are used for image matching. As the case study, the localization experiments of a mobile robot in an indoor environment are given. In these experiments, 3090 query images and 127 dataset images are used. This study includes five methods for feature detectors such as features from accelerated segment test (FAST), oriented FAST and rotated binary robust independent elementary features (BRIEF) (ORB), speeded-up robust features (SURF), scale invariant feature transform (SIFT), binary robust invariant scalable keypoints (BRISK), and five other methods for feature descriptors which are BRIEF, BRISK, SIFT, SURF, and ORB. These methods are used in 23 different combinations and it was possible to obtain meaningful and consistent comparison results using some performance criteria defined in this study. All of these methods are used independently and separately from each other as being feature detector or descriptor. The performance analysis shows the discriminative power of various combinations of detector and descriptor methods. The analysis is completed using five parameters such as (i) accuracy, (ii) time, (iii) angle difference between keypoints, (iv) number of correct matches, and (v) distance between correctly matched keypoints. In a range of 60°, covering five rotational pose points for our system, "FAST-SURF" combination gave the best results with the lowest distance and angle difference values and highest number of matched keypoints. The combination "SIFT-SURF" is obtained as the most accurate combination with 98.41% of correct classification rate. The fastest algorithm is achieved with "ORB-BRIEF" combination with a total running time 21303.30 seconds in order to match 560 images captured during the motion with 127 dataset images.

## 1. Introduction

Achieving beneficial information from visual sensors for indoor robots is crucial. In order to get this type of information capable of representing the real world with minimum loss of details, robots use various computer vision algorithms under the names of object detection, segmentation, and recognition. These algorithms work by matching and obtaining structural or inferred information about objects. Then, relating these separate low-level information sets, the algorithm constructs a framework in order to obtain semantic information. In this context, semantic information is necessary to get robots, machines, or digital systems to make sense from numerical data such as understanding what is happening in a scene or what the context of a speech/conversation could be. Consequently, semantic information requires more

*: Corresponding Author

computational effort and a deeper knowledge representation in comparison to low-level computing such as basic object detection and recognition algorithms.

Image matching has a wide range of applications in real world such as object and face recognition. In addition, image matching is the main operation for obtaining semantic information. On the other hand, image matching is still a challenging problem for real-time applications because of the amount of the data to be processed. One can briefly summarize the process of image matching as follows; 1) Constructing an appropriate feature database for the desired application, 2) Streaming live/recorded video or loading images to the system, 3) Computing the features of frame grabbed/captured from streamed live/recorded video or loaded images with the database, 4) Comparing the features, 5) Decision making about the quality of the matches such as accurately matched features.

Furthermore, our work includes design of an algorithm which may be applied to known indoor environments to obtain semantic information. In order to develop and test this algorithm, we created an image database (https://web.itu.edu.tr/bayraktare/Visual_Indoor_Dataset.rar) consisting 3090 images of an office environment and we specified various office objects within this database to match these objects with query images. We have selected 127 images which contain only one object in each image without any occlusions as database images to be compared by query images. In addition, these 127 images are taken as the ground-truth for our performance analysis. In order to accurately find the location of the robot, we consider the 3 height levels in a range of $30^0$. This means the localization algorithm may give six possible results for the same point. By using the output of the localization process, it is possible to determine the location of the robot within a cube which contains the 6 possible coordinates as boundary points. For this reason, scale and rotation performance of these methods are crucial for our localization algorithm.

This paper has been organized as follows; previous studies about feature detector-descriptor algorithms and performance evaluation of these methods are given in section 2, in 3rd section the datasets and the methods used in this study are explained, performance results are shown in 4th section, and finally, in section 5 we discuss the results.

## 2. Related Work

In literature, there is a wide range of studies based on feature detector-descriptor combinations of which some compare the feature detectors and feature descriptor methods; some of them argue the best detector-descriptor combinations, and a few of them focus on their performance in the recognition of objects. A feature in an image can be defined in a specific 2-dimensional structure that is composed of a detector and a descriptor. In this structure, the detector finds the repeatable interest points, and the descriptor is a distinctive specification that is obtained by computing each detected feature which can be matched between different images.

It is commonly accepted that SIFT [1], SURF [2], BRISK [3], ORB [6] methods used in this work consist of similar content and it may be given in 4 steps as follows; 1) Scale-space representation, 2) Key-point localization, 3) Orientation assignment, and 4) Key-point descriptor as given in [5]. In other words, one can summarize the first three of these steps as detector, and the last one as descriptor. Besides, some of these methods may include both the detector and descriptor, while some are distinctly known as detector or descriptor. As an example, FAST [7, 8] is a detector method and BRIEF [4] is a descriptor method. In the step of scale-space representation, SIFT applies a series of Difference-of-Gaussian (DoG) filters for multiple scales, and in this way we may get DoG filtered and down-sampled versions of the original image. SIFT descriptor is composed of a 4x4 array of gradient orientation histogram weighed by the gradient magnitude. On the other hand, scale-space representation in SURF is based on sums of 2D Haar wavelets using integral images that approximate Gaussian derivatives given by [2]. SURF detector approximates the determinant of Hessian matrix which will give a local maximum. SURF descriptor is a 64-dimensional vector which is obtained by summing the Haar Wavelet coefficients over 4x4 pixels around the key-point. As described in [7] and [8] FAST, a detection method that is actually used to detect corners, uses a circle consisting of 16 pixels around candidate corner pixels to classify whether that point is a corner or not by comparing these 16 pixels' brightness with the intensity of candidate pixel including a threshold. BRIEF is a binary descriptor based on pairwise intensity comparison. The detector part of the BRISK is given at [3] as computing FAST score, which is computed at each octave and intra-octave separately, across scale space and obtained continuous maximum across scales by calculating the sub-pixel maximum across patch at pixel level non-maximal suppression. BRISK descriptor contains concatenated brightness results tests with a binary string and it is rotation and scale invariant apart from BRIEF. Descriptor part of ORB is a method that is similar to BRIEF with rotation and scale invariances. Detector part of ORB applies FAST detector in a Gaussian pyramid.

The system proposed in [5] demonstrates the performance results of 4 descriptors [SIFT, SURF, BRISK, and FREAK] and looks for the best matching results in terms of detection accuracy and speed in the context of detecting the abandoned objects in real-time. This method is very sensitive to disturbances and the robustness is not completely ensured in the experimental setup. In addition to this, the camera system movement of the work is very limited such as linear back and forth motion. Another study, which investigates the detector and descriptor methods for photogrammetric applications [9], compares 5 interest point detectors with respect to correct detected corners, their localizations, the density of detected points/regions, but it is clear that the number of methods is very limited and performance analysis does not cover the time cost for these methods. Similar to a part of our work, [10] analyzes the different detector-descriptor combinations with 7 detectors and 2 descriptors with the aim of finding the best combination. They used a dataset that includes 60 scenes from 119 positions with 19 different illumination conditions. They conclude as a result of their experiments that DOG or MSER detector with a SIFT or DAISY descriptor as the best combination by using a performance measure computed from area under receiver operating characteristic (ROC) curve which depends on a proportion of the distance values between the best matched features.

As a matter of course, previous studies have used different measuring scales and criteria in order to evaluate the performance outputs of different feature detector-descriptor combinations. In [11], although it does not consider time performance, the performance evaluation is realized counting the number of correctly detected interest points and their locations with comparing the density and relative orientations of these points with stereo pairs for 5 interest point detectors (Förstner, Heitger, Susan, Harris, and Hessian) and two region detectors/descriptors (Harris-Affine and SIFT). Another study, [12], considers the occlusions and realized using a moving camera gives the performance comparison results for 4 descriptors (SURF, SIFT, BRISK, and FREAK). This paper uses accuracy and speed as performance criteria, but it is very sensitive to disturbances, and the robustness is provided by limiting the movement of the robot just by a linear back-and-forth motion at the experimental setup. Although it tells that the real-time detection, normalized cross correlation, which scans all the frame which makes this method slow rotation-variant, is applied to the binary image that means loss of information in the image used as image comparison method. [13] compares only binary descriptors (SIFT, FAST, BRIEF, BRISK, ORB, and FREAK) and their combinations without giving any details about these combinations' total performance and compliance of different detectors and descriptors. The performance evaluation results are obtained by matching the given images and pixel based distance values of the corresponding points. In this study, SIFT is assumed to be ground-truth, but ground-truth would be obtained by direct matching of images more precisely. Another local descriptor comparison is given in [14] for only 4 methods. These 4 methods are based on Harris-Affine detector and they are compared with regard to the complexity of the compared methods' individual parameters and usage areas with detection rate with respect to false positive rate as evaluation criterion based on the calculation of the area under Receiver Operating Characteristics (ROC) curves. In addition to the comparison of local detectors and descriptors, [15] investigates the performance of in terms of two properties; robustness and distinctiveness using a unified framework. The framework is composed of two steps; the first step is detector evaluation criteria, which takes into account the accuracy of localization under different conditions, and the repeatability score for 6 detectors (Harris-Affine, Hessian-Affine, Maximally stable extremal regions, Intensity based regions, Geometry based regions, Salient regions). The second step is descriptor evaluation criterion that considers the distinctiveness, which is measured by ROC of the number of correct matches with respect to the number of corresponding regions against false positive rate with a distance threshold for 6 descriptors (SIFT, steerable filters, differential invariants, complex filters, moment invariants, cross-correlation). On the other hand, [16] gives the effects of different detectors (SIFT, SURF, BRISK, ORB, FAST, GFTT, STAR) and descriptors (SIFT, SURF, BRISK, ORB, BRIEF, FREAK) on RGB-D SLAM (Simultaneous Localization and Mapping) methods. The performance evaluations of these methods are investigated in terms of accuracy by measuring empirically the distance of matched objects and time to process per frame. The detector-descriptor combinations of this study are limited and only BRIEF and FREAK descriptors are combined with other detectors. In an addition to the comparison of feature detectors and descriptors performance on visual SLAM, [18] compares the effects of 6 feature descriptors (BRIEF, BRISK, FREAK, ORB, SIFT, SURF) on graph-based VSLAM algorithm according to localization accuracy and motion speed of the camera for real-time performance using two different datasets as being used as two different motion scenario. Even though, both the number of performance parameters and compared descriptors are limited, and there are no detectors compared, similarly to our accuracy and time performance evaluation results given in Table 1, the results are clearly show that the accuracy of SIFT and the speed of BRIEF are the best options. In the context of image search and fine-grained classification, [17] provides the performance assessment results in terms of accuracy and time by proposing new approaches such as modifying Harris, Hessian and DoG detectors to extract dense patches, changing the scale of selecting these patches' edges, and filter selection is made by their own method to locate patches. Although this paper studies on the developing the accuracy and process time, it is underlined that not to target high repeatability. Besides, [19] examines the JPEG compression's effects on different feature detector-descriptor combinations. They used VLBenchmarks [21] framework to test their 60 combinations composed of 10 detectors with 6 descriptors, which the dependency to this framework may harm some important properties about performance evaluation results such as repeatability, accuracy, speed and extracted feature numbers. Performance evaluation parameter of this study is very limited with mAP (mean average precision) Scores, which is computed by taking the average of 55 same-level compressed query images with JPEG with a quality range from 4 to 20 after applying 3 deblocking methods (spatial domain, frequency domain, hybrid filtering), by comparing the number of feature detector-descriptor combinations and the feature extraction speed is not considered as a performance evaluation criterion.

Similar to [15], [20] gives the comparison of performance evaluation results about affine covariant region detectors using a structured and textured scenes under different conditions such as viewpoints changes, scale changes, illumination and blurring variations. This comprehensive study assesses the performance in terms of measuring the repeatability by the detector's performance on determining the corresponding scene region, and the accuracy with respect to regions' shape, scale, and localization. In addition, also, the distinctiveness of the detected regions is evaluated as another parameter for performance to emphasize the discriminative power of these methods.

Our study is one of the most comprehensive comparisons available because we give 23 different feature detector-descriptor performance evaluation results in terms of accuracy and speed in a real-world scenario including localization experiment. Additionally, we present 19 different feature detector-descriptor comparisons regarding the number of correct matches, mean angle difference between key-points, and minimum distance metrics. Although, as mentioned above, the literature studies mostly focused on accuracy and time performance evaluations, this work investigates the time consumed to match two features, time performance from start to finish of the comparison of all query images with all dataset images for different detector-descriptor methods combinations, accuracy values of these combinations, and the relative relations of different detector-descriptor combinations with respect to correct matches, mean angle difference between key-points, and minimum distance between key-points.

## 3. Data and Methodology

In this section we give details about the dataset and query images which we created for this study. Query image dataset is created by grabbing indoor images with the dimensions of 555x480 pixels using a low-cost CMOS camera from a laboratory environment by rotating the camera by 15$^o$ with an apparatus at predetermined points for 3 different heights. Besides, the dataset that contains the template images to be matched with the query images which are chosen from the query image dataset. Figure 1.a shows the image grabbing process and the hypothetical volumetric location cube, and the objects in the template images are given in Figure 1. b. A collage of query images grabbed from two different points is displayed in Figure 1. c.

In addition, before obtaining performance results of a combination of methods between query and template images, two elimination steps are evaluated in order to prevent obtaining trivial results. After a satisfactory matching score is obtained then the robot location is determined approximately in a hypothetical cube.

In the first step, we used FAST method to detect the key-points of the query image and determined a hysteresis threshold that provides us to eliminate images which have less key-points than the lower threshold or have much key-points than the upper threshold. In the second step, we used histogram comparison results to eliminate query images before matching process. Thus we have same number of matching results for all combinations. After these steps we achieve the performance results at matching process of the algorithm.

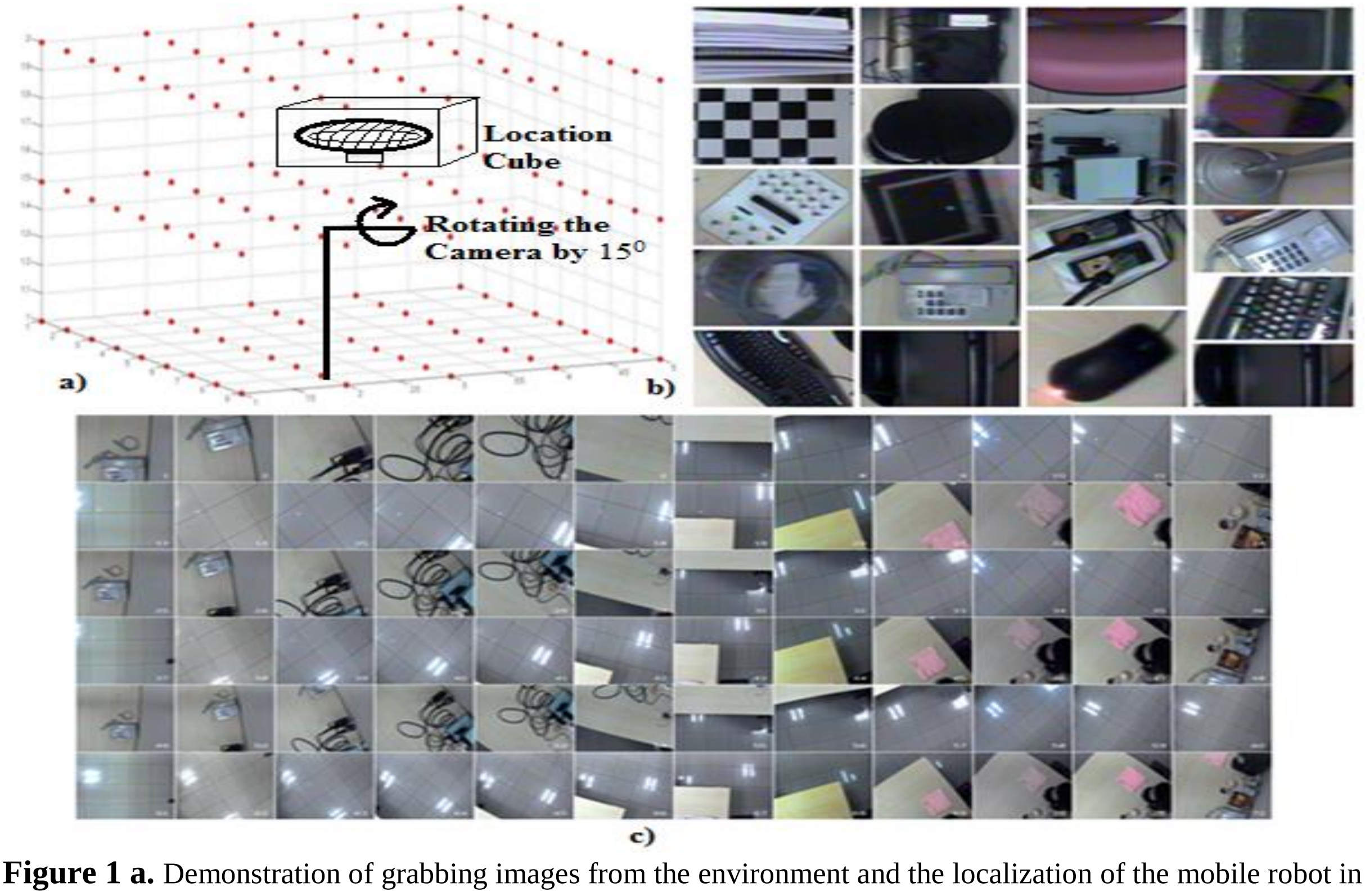

**Figure 1 a.** Demonstration of grabbing images from the environment and the localization of the mobile robot in a hypothetical location cube. **b.** A part of template image dataset. **c.** A sequence of query image dataset which grabbed from 2 points.

In order to visualize experimental localization results, two different tools are used. Figure 2.a demonstrates the mobile robot’s path during its flight in the office room and Figure 2.b displays the same path in a virtual reality environment. The red symbols indicate the recognized location of the robot which means a matched query image with a template image and it can be seen from the Figure 1. a that there are much more recognized points at the last stage of the movement than the beginning because of the speed of the robot is faster in the beginning and then slows down.

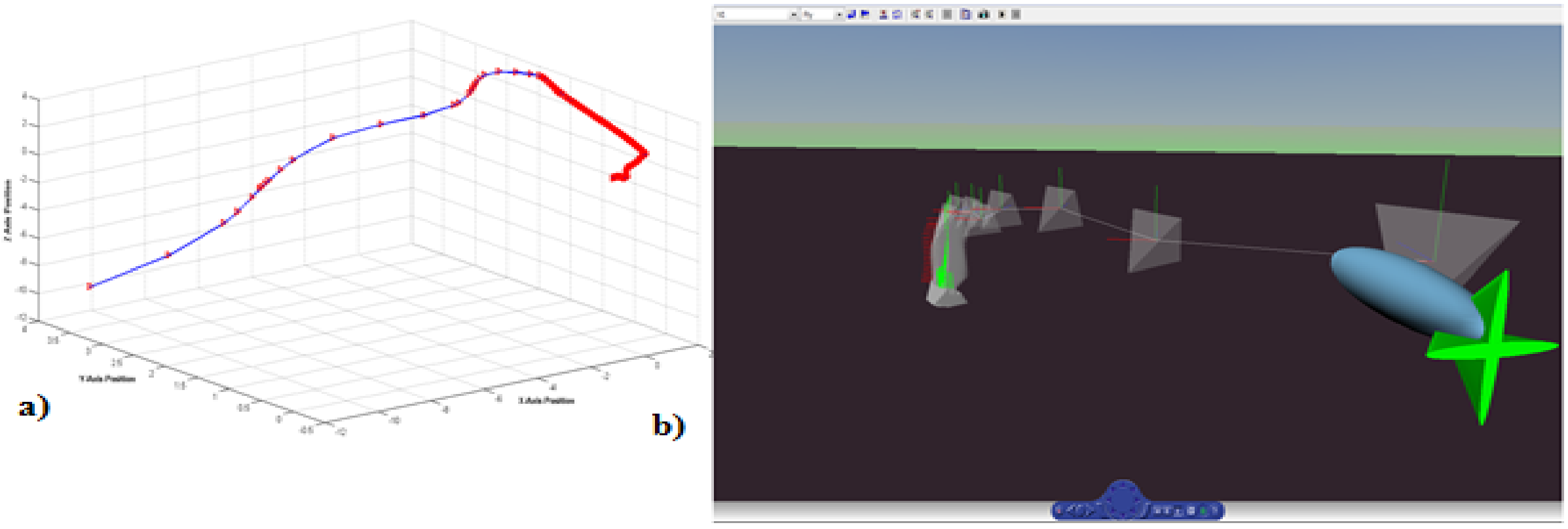

**Figure 2** Localization experiment results. **a.** Demonstration of the location of the mobile robot in 3D coordinates. **b.** Path followed by mobile robot using a different visualization tool.

## 4. Performance Analysis

This section demonstrates the performance results using different parameters such as time for total computation, number of matches per second, and accuracy given in Table 1 and mean key-point angle differences, number of correct of matches, distance metrics of matches, given in Figure 3.

In order to complete Table 1, we compared all query images grabbed from the office room with template images chosen from query images. Thus, we observed the parameter changes whether an image pair (query and template) is related according to its position/rotation or these query and template images are irrelevant. Before the comparison section of the program, we eliminate some of the query images using a hysteresis threshold with respect to the number of key-points which are computed using FAST method. The first column of Table 1, total running time of the algorithm means the time passed until the robot finishes the same number of image comparison process using different combinations on the same path between 560 query images and 127 template images. The last column is the time passed to match correctly per features extracted by different feature-detector-descriptor combinations. It is clear from Table 1 that the fastest combination consists of *ORB* for both key-point detector and key-point descriptor. On the other hand, minimum number of correct matches per second belongs to the combination of *BRISKSIFT*; conversely, maximum number of correct matches per second belongs to the combination of *FASTBRIEF*. Furthermore, *SURFSIFT* combination has the biggest running time. In an addition, if the SIFT is used as the key-point descriptor, time for total computation is lower than all the other combinations for each key-point detector methods. It can be easily estimated that the accuracy for the 127 template images are %100 for all combinations because the template images are chosen from the query images and these are exactly the same.

$$\%Accuracy = \frac{\sum(True\ poisitives + True\ negatives)}{\sum(Total\ Cases)} * 100 = \frac{127+0}{127} * 100 = \%100$$

Therefore, giving these accuracy rates is trivial. However, we formulated a different accuracy term that includes the 3 height level for 5 different pose angles from -$30^0$ to $30^0$ according to the current pose angle and height, and if the histogram comparison value is higher than the threshold value (0.9) of that comparison result, it is accepted that this comparison is made for the same point at different pose angles. This measure of accuracy works because the algorithm is designed so that if any match is achieved in this range in terms of given parameters, the robot location detected at that point is acceptable. Furthermore, in Table 1 the accuracy rates are given for explained situation and *SIFTSURF* is ascertained as the most accurate combination with % 98.41. Moreover, as one can see that excluding BRISK method as key-point detector combinations, the best accuracy rates are achieved when the SURF method is the key-point descriptor.

**Table 1** Performance Results for All Combinations.

| KEYPOINT DETECTOR TYPE | Parameters Utilized in Performance Analysis | | | | |
|---|---|---|---|---|---|
| | KEYPOINT DESCRIPTOR TYPE | Total Time of the Algorithm (sec) | Accuracy | Ground-Truth | Number of Correct Matches per Second |
| **ORB** | BRIEF | 21303.299 | %62.83 | 127*3*5 | 1457.011 |
| | BRISK | 23461.675 | %74.28 | 127*3*5 | 956.158 |
| | SIFT | 97603.701 | %72.28 | 127*3*5 | 318.012 |
| | SURF | 79391.059 | %97.90 | 127*3*5 | 390.965 |
| | ORB | 21330.015 | %63.62 | 127*3*5 | 1455.186 |
| **SURF** | BRIEF | 32277.702 | %62.36 | 127*3*5 | 1568.788 |
| | BRISK | 35133.182 | %63.52 | 127*3*5 | 1275.433 |
| | SIFT | 196487.667 | %68.82 | 127*3*5 | 280.432 |
| | SURF | 79135.063 | %89.54 | 127*3*5 | 696.295 |
| | ORB | 35074.987 | %63.78 | 127*3*5 | 1414.503 |
| **SIFT** | BRIEF | 30938.270 | %62.73 | 127*3*5 | 879.923 |
| | BRISK | 32422.058 | %64.67 | 127*3*5 | 920.443 |
| | SIFT | 45919.045 | %62.31 | 127*3*5 | 698.415 |
| | SURF | 35319.080 | %98.41 | 127*3*5 | 908.024 |
| **FAST** | BRIEF | 23355.701 | %62.52 | 127*3*5 | 2736.879 |
| | BRISK | 33752.854 | %63.20 | 127*3*5 | 454.341 |
| | SIFT | 56154.363 | %72.44 | 127*3*5 | 1373.775 |
| | SURF | 37357.531 | %88.30 | 127*3*5 | 2065.004 |
| | ORB | 22734.255 | %62.62 | 127*3*5 | 275.125 |
| **BRISK** | BRIEF | 20517.530 | %64.62 | 127*3*5 | 320.6933 |
| | SIFT | 34284.954 | %86.61 | 127*3*5 | 202.3933 |
| | SURF | 26043.988 | %80.32 | 127*3*5 | 266.4356 |
| | ORB | 23335.765 | %69.76 | 127*3*5 | 289.8432 |

In order to make our comparison more comprehensive, we give the performance results for different pose cases using the query and template image matches within an rotational pose range of $[-30^0, 30^0]$ for 5 cases in Figure 3. From Figure 3.a, it is clear that the all methods are scattered in a wide range in 3-dimensional space, and also *FASTSURF* combination gives the best results for all rotations excluding the comparison given for the same images with respect to number of correct matches, mean of angle of differences between correctly matched key-points and minimum distance metrics. If we limit our comparison parameters to two of recent parameters for these comparison results by taking the application priorities into account, the best combination may change. For instance, if the minimum distance metrics and the number of correct matches are important for an application, then for $15^0$ and $30^0$ there are 4 best results (SURFSURF, SURFBRIEF, FASTORB, and SURFBRISK). It is possible to make such analysis for any kind of priorities from the given performance evaluation parameters.

If the comparison is given for the same image, the combinations align on a straight line at different points. In this case, *SURFSURF* and *SURFSIFT* combinations both give the best results, and *SIFTBRIEF*, *ORBBRISK*, *SIFTSURF*, *SIFTBRISK*, and *SIFTSIFT* are very close to each other giving the worst results. As a matter of course, only the number of correct matches change between 220 matches and 990 matches, other parameters are the same for all combinations.

Whether the number of correct matches is changing between around 200 and 1000 for all rotation cases, minimum distance metrics and mean of angle difference values between key-points vary. For positive signed rotations ($15^0$, $30^0$), minimum distance metrics values change between 0 and 170 pixels. On the other hand, for negative signed rotations ($-15^0$, $-30^0$), minimum distance metrics values change between 0 and 300 pixels. Furthermore, average of angle difference values between correctly matched key-points vary from $-30^0$ to $30^0$ for negative signed rotations, and from $-10^0$ to $40^0$ for positive signed rotations.

Consequently, we may obtain the worst combination results with respect to present parameters by calculating the geometrical distance using the relative geometric positions of all combinations from the figures to the best combination. For the rotations of $30^0$, $15^0$, and $-30^0$, *ORBSIFT*, for the same image comparison *SIFTBRIEF*, and for $-15^0$, *SURFSIFT* combinations are obtained as the worst combinations with respect to given comparison parameters.

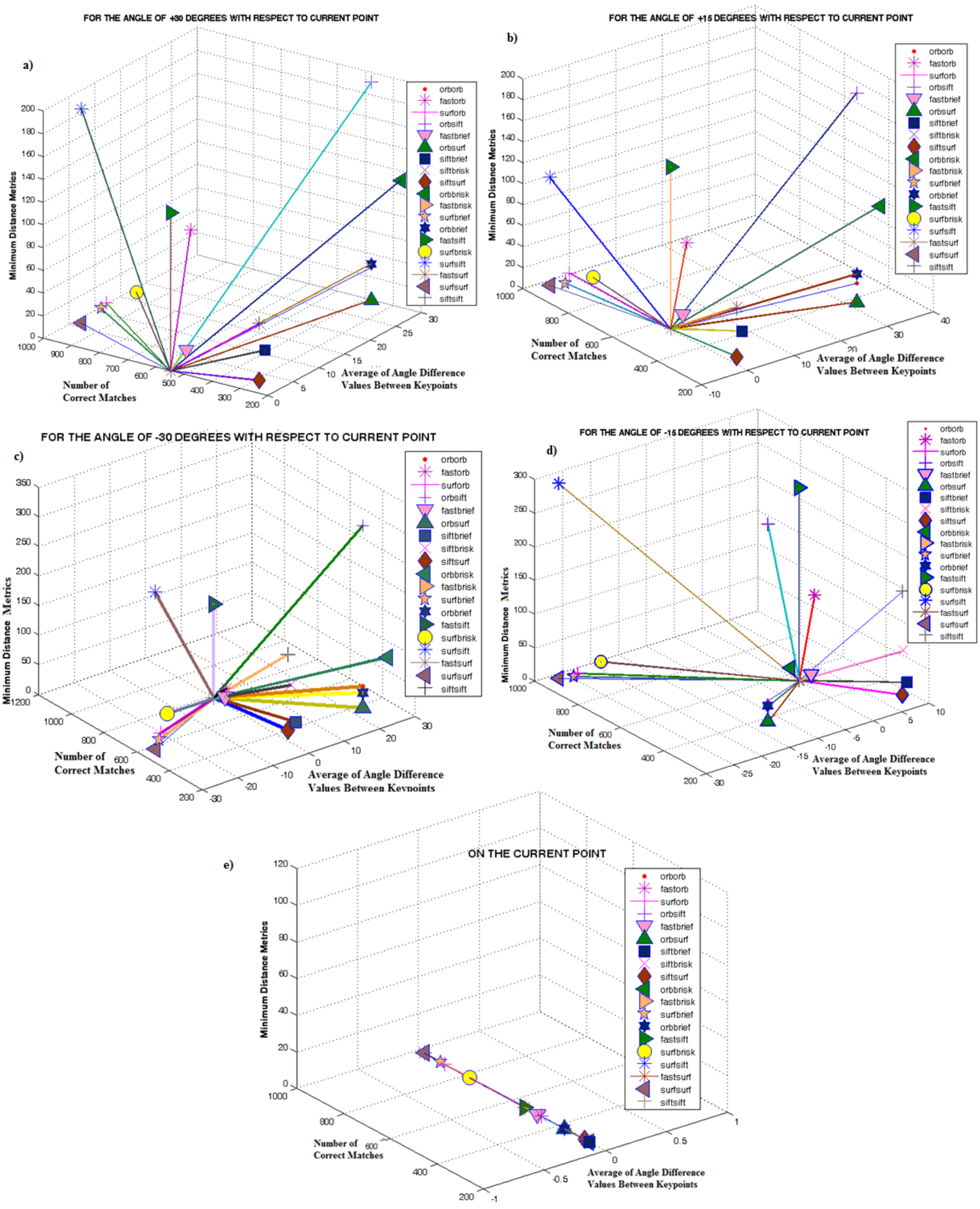

**Figure 3** Performance results for the different angle cases. These comparison results are given for the points that are rotated from $-30^0$ to $+30^0$ according to the current point which query image and template image is the same with respect to number of correct matches, minimum distance metrics and the average of angle difference values between key-points. Lower the average of angle difference values between key-points, and minimum distance metrics, and higher the number of correct matches is desired for the best result. In summary for all subfigures displayed in here is the longer to the central point, which indicates the best method, the worse the performance. **a.** Comparison of the template image with $+30^0$ rotated query image from the same environment and height. **b.** Comparison of the template image with $+15^0$ rotated query image from the same environment and height **c.** Comparison of the template image with $-30^0$ rotated query image from the same environment and height **d.** Comparison of the template image with $-15^0$ rotated query image from the same environment and height **e.** Comparison of the template image with the same image as query image.

## 5. Conclusion and Future Work

Developing good feature detectors and descriptors is still a challenging research topic. Independent from the development of hardware, especially for embedded and/or onboard applications, and in any autonomous system, the parameters such as computation time, robustness, repeatability, and accuracy are crucial in the context of soft algorithm development of feature detectors and descriptors. It has been a usual path to create a new feature detector-descriptor method by combining previous detector and descriptor methods in an effective algorithm. Our study provides more comprehensive information with more performance evaluation parameters (in terms of accuracy and temporal costs such as the running time from start to end, time per matched features) wider coverage of methods (i.e. 23 combinations). In addition to these commonly used metrics, we give the performance results of 19 combinations with respect to new metrics, such as mean keypoint angle differences, number of correct matches, and the distance metrics of matches.

It is clear from our results that there are trade-offs between different parameters and performance criteria when different feature detector-descriptor combinations are analyzed. If a wide rotation range is desired to be matched, then the algorithm finds weak features and matches are not realized because of predefined threshold values. On the other hand, if a narrow rotation range is desired, the algorithm finds too many features to match and this causes the increase at total loop time because of an increase at number of correct matches per unit time. The lowest total running time belongs to the combination of BRISK-BRIEF with 64.62% accuracy rate and 320.6933 correct matches per second. The highest running time is for the SURF-SIFT combination with 68.82% accuracy rate and 280.432 correct matches per second. Additionally, SIFT-SURF combination has 98.41% accuracy rate in a total of 35319.080 seconds with 908.024 correct matches per second. Furthermore, FAST-SURF combination is the method to give the best results for the angular rotations of 30°, 15°, -15°, and -30° for comparisons. Moreover, for the case of the comparison of the same images, SURF-SURF and SURF-SIFT combinations give the best results with respect to number of correct matches, mean of angle of differences between correctly matched keypoints and minimum distance metrics.

Future research in the computer vision, perception and robotics areas can benefit from the results provided in this study. From the investigated methods, the individual priorities of different applications can easily be reflected in combinations as the best option. To be more specific, our results can improve the applications in the fields of object recognition by image/object matching in different conditions, and in visual SLAM field by extracting robust features with its generic outcomes.

In our future work, using the results of this study, an experiment in real-time on our ongoing humanoid project UMAY [21] with a wider object database will be performed to get the semantic information of unknown indoor environments. Since the recognition of the environment is crucial for the robot in terms of executing or interpreting the given tasks in a dynamical way, object recognition algorithms here can provide an optimized method to start with processing the visual information. As a consequence, this framework can be considered to form a basis for future applications involving extracting of visual semantic cues.

## Acknowledgements

The research was conducted at the Mechatronics Research and Education Center at Istanbul Technical University.